\documentclass[10pt]{IEEEtran}
\IEEEoverridecommandlockouts
\usepackage{cite}
\usepackage{amsmath,amssymb,amsfonts}
\usepackage[linesnumbered,ruled,vlined]{algorithm2e} 
\usepackage{graphicx}
\usepackage{xcolor}
\usepackage{booktabs}
\usepackage[top=1in, bottom=0.8in, left=0.75in, right=0.75in]{geometry}
\usepackage[hidelinks]{hyperref}
\begin{document}
\title{{Energy-Efficient Autonomous Aerial Navigation with Dynamic Vision Sensors: A Physics-Guided Neuromorphic Approach}}

\author{
    Sourav Sanyal\IEEEauthorrefmark{1}\thanks{\IEEEauthorrefmark{1}These authors contributed equally to this work.}, 
    Amogh Joshi\IEEEauthorrefmark{1}, 
    Manish Nagaraj, 
    Rohan Kumar Manna, 
    Kaushik Roy \\
    \IEEEauthorblockA{School of Electrical and Computer Engineering, Purdue University}
}

\maketitle

\begin{abstract}

Vision-based object tracking is a critical component for achieving autonomous aerial navigation, particularly for obstacle avoidance. Neuromorphic Dynamic Vision Sensors (DVS) or event cameras, inspired by biological vision, offer a promising alternative to conventional frame-based cameras. These cameras can detect changes in intensity asynchronously, even in challenging lighting conditions, with a high dynamic range and resistance to motion blur. Spiking neural networks (SNNs) are increasingly used to process these event-based signals efficiently and asynchronously. Meanwhile, physics-based artificial intelligence (AI) provides a means to incorporate system-level knowledge into neural networks via physical modeling. This enhances robustness, energy efficiency, and provides symbolic explainability. In this work, we present a neuromorphic navigation framework for autonomous drone navigation. The focus is on detecting and navigating through moving gates while avoiding collisions. We use event cameras for detecting moving objects through a shallow SNN architecture in an unsupervised manner. This is combined with a lightweight energy-aware physics-guided neural network (PgNN) trained with depth inputs to predict optimal flight times, generating near-minimum energy paths. The system is implemented in the Gazebo simulator and integrates a sensor-fused vision-to-planning neuro-symbolic framework built with the Robot Operating System (ROS) middleware. This work highlights the future potential of integrating event-based vision with physics-guided planning for energy-efficient autonomous navigation, particularly for low-latency decision-making. 



\end{abstract}
\thispagestyle{empty}

\section{Introduction}
"How can robots perceive and react to their environments as efficiently as humans?" This question lies at the heart of autonomous navigation research, driving efforts to create systems that are both adaptive and efficient. Autonomous navigation has been a hot topic of robotics research, with a growing emphasis on leveraging artificial intelligence (AI) to enhance performance. However, traditional AI-based navigation algorithms often fall short in dynamic and reactive scenarios, such as close-in obstacle avoidance, where rapid decision-making is paramount. These limitations stem from the inability of conventional sensing modalities to provide data with sufficient speed and reliability. Vision-based algorithms, while effective for high-level decision-making, face challenges such as high energy consumption and latency inherent to frame-based cameras, undermining their utility in energy-constrained aerial systems.

Neuromorphic event cameras, also known as Dynamic Vision Sensors (DVS) \cite{dvs1,dvs2,dvs3}, have emerged as a promising alternative to frame-based cameras. Inspired by biological vision, these sensors detect changes in intensity asynchronously, offering advantages like low latency, high dynamic range, and robustness to motion blur. Unlike conventional cameras that capture redundant static frames, event cameras record only significant intensity changes, reducing bandwidth and enabling data capture at microsecond granularity. This asynchronous data stream aligns naturally with the computational model of spiking neural networks (SNNs) \cite{lif,lee2020spike}, which process information in a similarly event-driven manner. Works in \cite{ref1, ref2, ref3} have demonstrated the applicability of SNNs in real-time control and decoding of sensorimotor tasks, particularly in aerial robotics and insect-scale neuromotor systems. Although these studies highlight the effectiveness of SNNs in neuromorphic control and decoding, they do not directly address the integration of event-based vision for perception and planning. In this work, we focus on harnessing the asynchronous visual signals from DVS sensors to guide real-time robot behavior, bridging the gap between neuromorphic perception and control.

Simultaneously, physics-based artificial intelligence (AI) has gained traction in robotics, providing a means to integrate system-level physical knowledge into neural networks. By embedding physical models into the learning process \cite{raissi2019physics,karniadakis2021physics}, physics-based AI enhances model robustness, interpretability, and energy efficiency. In the context of aerial robotics, where energy efficiency and real-time responsiveness are critical, such integration becomes highly advantageous. Physics-guided neural networks (PgNNs) leverage prior knowledge of system dynamics \cite{NICODEMUS2022331,rampnet,chee2022knode} to predict optimal trajectories, reducing reliance on extensive training datasets and computational resources.

\begin{figure*}[!t]
  \centering
  \includegraphics[width=\textwidth]{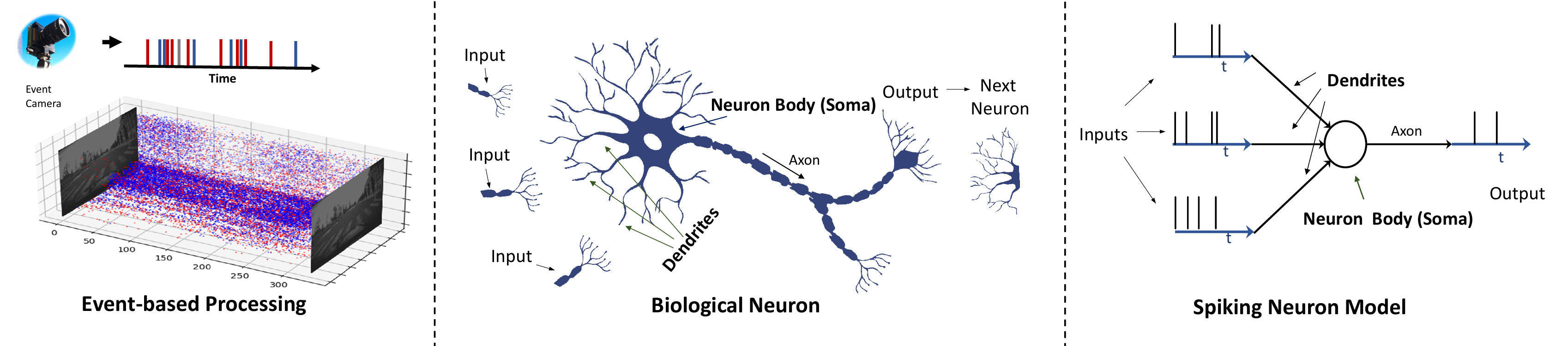}
  \caption{%
    \textbf{Event-Based Vision and Spiking Neuron Model:}
    Illustration of the conceptual flow from event-based sensing to biological inspiration
    and spiking neural networks (SNNs). 
    \textit{Left:} An event camera outputs discrete intensity changes over time (red/blue dots) 
    rather than continuous image frames.
    \textit{Center:} Biological neurons communicate via discrete spikes, with dendrites 
    receiving inputs that the soma integrates before generating an output spike. 
    \textit{Right:} In SNNs, inputs are represented as spikes over time, and the neuron body (soma) 
    converts these event-driven signals into an output spike train, resembling biological neuronal firing.
  }
  \label{fig:event_based_processing}
\end{figure*}

In this work, we present an energy-efficient navigation framework that combines event-based vision with physics-guided planning for autonomous drones.  The system employs a neuromorphic event camera for detecting dynamic objects using a shallow SNN architecture in an unsupervised manner. Additionally, a lightweight PgNN, trained with depth sensor inputs, predicts near-optimal flight times, generating energy-efficient trajectories. By fusing depth and event-based data, the framework achieves robust object tracking and trajectory planning in complex environments.

The proposed framework is implemented in the Gazebo simulator  utilizing the Robot Operating System (ROS) middleware. Our system demonstrates the ability to detect and navigate through moving gates, avoiding collisions while minimizing energy consumption.  This work illustrates the potential of integrating neuromorphic event-based vision with physics-guided planning for aerial robotics.

\section{Preliminaries}
\subsection{\textbf{Neuromorphic Event Cameras}}

Neuromorphic event cameras, such as Dynamic Vision Sensors (DVS) \cite{dvs1,dvs2,dvs3}, represent a fundamentally different sensing modality compared to conventional frame-based cameras. Instead of capturing entire frames at fixed intervals, event cameras respond asynchronously at each pixel whenever a significant brightness change is detected. These changes trigger ``events,'' each recorded with the pixel coordinates, timestamp, and polarity. Formally, an event $e_k$ is generated at pixel $(x_k, y_k)$ if
\begin{equation}
  \label{eq:eventcamera}
  \log\bigl(I(x_k,y_k,t_k)\bigr) \;-\; \log\bigl(I(x_k,y_k,t_{\text{prev}})\bigr)
  \;\;\geq\; C,
\end{equation}
where $I(\cdot)$ is the brightness (or intensity) at that pixel, $t_{\text{prev}}$ is the last time an event occurred at $(x_k, y_k)$, and $C$ is a user-defined contrast threshold. An individual event can be expressed as a tuple
\[
e_k \;=\; \bigl(x_k,\, y_k,\, t_k,\, p_k\bigr),
\]
where $p_k \in \{+1, -1\}$ denotes the polarity (i.e., whether the intensity increased or decreased). Advantages of Event Cameras include:
\begin{itemize}
  \item \textbf{High Temporal Resolution}: Events are reported at microsecond granularity, supporting fast maneuvers.
  \item \textbf{Low Latency}: Data is streamed continuously and asynchronously, allowing near-instantaneous reactions.
  \item \textbf{Reduced Redundancy}: Only intensity changes are recorded, lowering bandwidth and data-processing demands.
  \item \textbf{High Dynamic Range \& Robustness}: They handle challenging lighting conditions and are minimaly sensitive to motion blur.
\end{itemize}

\subsection{\textbf{Spiking Neural Networks (SNNs)}}

Spiking Neural Networks (SNNs) \cite{lif,lee2020spike} emulate the event-driven communication seen in biological neurons, making them a natural fit for the asynchronous output of neuromorphic event cameras. Unlike traditional artificial neural networks, SNNs process information through discrete ``spikes'' rather than continuous activations.\\

\noindent{\textbf{LIF Neuron Model}}:
A widely used spiking neuron model is the Leaky Integrate-and-Fire (LIF) neuron. Its membrane potential $V(t)$ evolves according to
\begin{equation}
  \tau_m \,\frac{dV(t)}{dt} \;=\; -\,V(t) \;+\; R\,I(t),
\end{equation}
where:
\begin{itemize}
  \item $\tau_m$ is the membrane time constant,
  \item $R$ is the membrane resistance,
  \item $I(t)$ is the synaptic input current.
\end{itemize}
A spike is generated whenever $V(t)$ exceeds a threshold $V_{\text{th}}$, at which point $V(t)$ is reset to $V_{\text{reset}}$. Mathematically,
\[
V(t) \;\xrightarrow{V(t)\,\ge\,V_{\text{th}}}\; V_{\text{reset}},
\quad
\text{(spike event generated)}.
\]
The event-driven mechanism of SNNs significantly reduces energy consumption, since computation primarily occurs when spikes are present. We will use these principles to design our perception front end (Section \ref{subsec:neuromorphic_detection}).

\begin{figure*}[!t]
    \centering
    \includegraphics[width=1\textwidth]{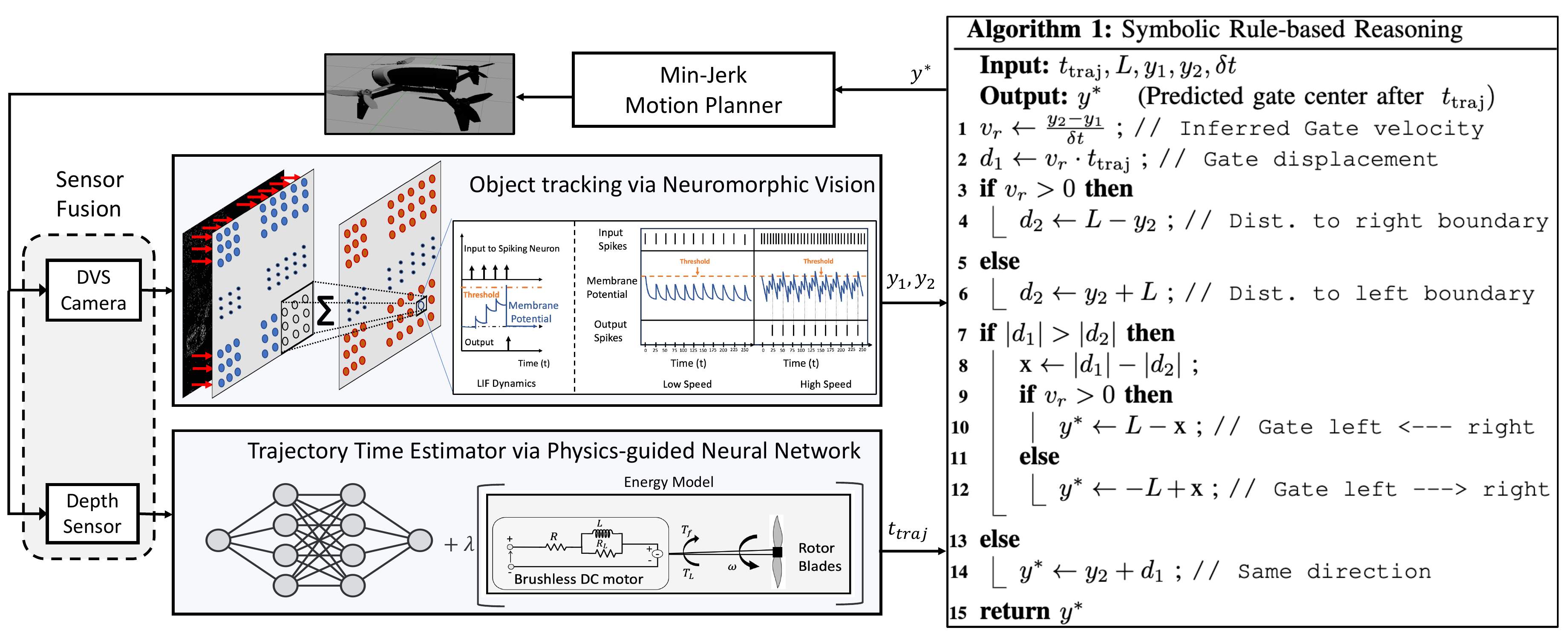}
    \caption{%
        \textbf{System architecture:} Integrating neuromorphic vision, 
        physics-guided neural networks, and symbolic rule-based reasoning. 
        Event and depth streams feed the SNN and PgNN, informing real-time 
        navigation decisions. Adapted from \cite{evplanner}. The motion planner is based on the work in \cite{mellinger2011minimum}.
    }
    \label{fig:method}
\end{figure*}

\subsection{\textbf{Physics-Guided Neural Networks (PgNNs)}}

Physics-Guided Neural Networks (PgNNs) \cite{raissi2019physics,karniadakis2021physics,NICODEMUS2022331,rampnet,chee2022knode} integrate system dynamics or physical constraints directly into the network training process. This approach improves interpretability, robustness, and data efficiency, which are crucial attributes for aerial robotics.

\noindent{\textbf{Embedding Physical Laws}}:
Consider a simplified dynamic model for a drone, where $\mathbf{x}(t)$ is position, $\mathbf{v}(t)$ is velocity, and $m$ is the mass:
\begin{equation}
\label{eq:dynamics}
  \dot{\mathbf{x}}(t) \;=\; \mathbf{v}(t), 
  \quad
  \dot{\mathbf{v}}(t) \;=\; \frac{1}{m}\bigl(\mathbf{F}_{\text{thrust}}(t) \;-\; m\,\mathbf{g}\bigr)
  \;-\;\mathbf{d}\bigl(\mathbf{v}(t)\bigr),
\end{equation}
where $\mathbf{F}_{\text{thrust}}(t)$ is the thrust force, $\mathbf{g}$ is gravitational acceleration, and $\mathbf{d}(\cdot)$ models drag or other aerodynamic effects. A PgNN can be designed such that a portion of its outputs obey (or approximate) these dynamics, restricting the solution space to physically feasible behaviors. This is one possible example. Later, we will see how we can apply this principle for energy models as well (Section \ref{subsec:physics_guided}).

\section{Proposed Approach}
\label{sec:proposed_approach}

We adopt neuromorphic computing concepts to process event-based data streams. By leveraging spiking neural networks (SNNs), we aim to emulate the inherent efficiency of biological neurons, which communicate primarily through discrete spikes rather than continuous signals.


Building on neuromorphic principles, our approach harnesses an event camera for real-time object detection and tracking. Subsequent modules—including a spiking neural network (SNN) block, a physics-guided neural network (PgNN) for energy-optimal trajectories, and a rule-based planner—operate on sparse event data to generate collision-free flight paths.
\begin{figure*}[t]
    \centering
    \includegraphics[width=1\textwidth]{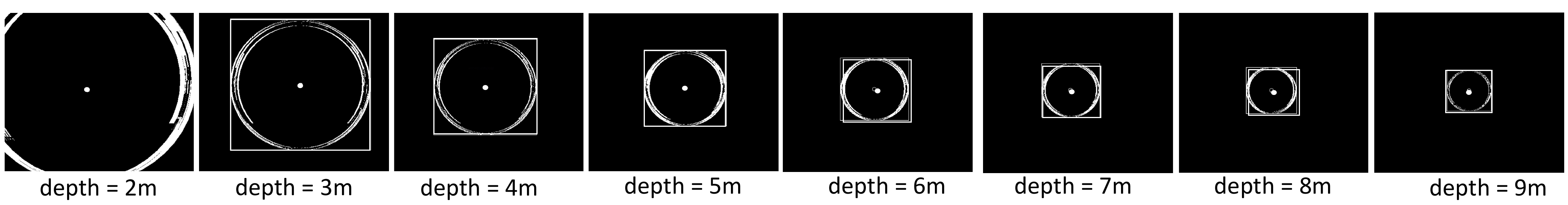}
    \caption{%
        \textbf{Event-based Object Detection at Various Depths.}
        Each sub-panel shows neuromorphic event output and a bounding box 
        around a moving gate. Although event density decreases
        with increasing depth, the SNN continues to isolate and track the gate
        in real time.
    }
    \label{fig:ev_box}
\end{figure*}
We propose a system that processes sparse, event-based sensor data to generate collision-free trajectories, as illustrated in Figure~\ref{fig:method}. Our approach integrates:
\begin{enumerate}
    \item A \textbf{spiking neural network (SNN)} for event-based object detection,
    \item A \textbf{physics-guided neural network (PgNN)} for near-minimum-energy destination prediction, and
    \item A \textbf{rule-based planner} for handling moving obstacles.
\end{enumerate}
All components interact in a \textbf{ROS} and \textbf{Gazebo} simulation environment: the SNN publishes bounding-box coordinates of the moving gate, the PgNN outputs an estimated flight time, and a symbolic planner node fuses both to generate velocity commands for the drone’s low-level controller.

\subsection{\textbf{Neuromorphic Vision-based Object Detection}}
\label{subsec:neuromorphic_detection}

In event-based cameras, each pixel independently fires upon detecting changes in brightness, producing sparse asynchronous data streams. Unlike frame-based cameras that capture images at fixed intervals, event-based cameras generate events only when a change in the scene occurs, making them highly efficient for capturing fast motions and dynamic environments. This event-driven nature allows them to have a high temporal resolution and low latency, making them particularly advantageous for real-time applications.

Inspired by \cite{nagaraj2023dotie}, we adopt a biologically-plausible approach using leaky integrate-and-fire (LIF) neurons to detect fast-moving objects. The LIF neuron model simulates how biological neurons process incoming stimuli by accumulating membrane potential over time. Specifically, the membrane potential $V[t]$ of an LIF neuron evolves as:

\begin{equation}
V[t] = \beta , V[t_{n-1}] + W , X[t],
\label{lif_eq}
\end{equation}

where $\beta$ is the leak factor representing the decay of the membrane potential, $W$ is a learnable weight matrix that scales the contribution of the input, and $X[t]$ is the aggregated spike input at time $t$. The leak factor $\beta$ controls how quickly the neuron forgets previous inputs, with smaller values corresponding to faster decay.

Whenever the membrane potential $V[t]$ exceeds a predefined threshold $V_{\mathrm{th}}$, the neuron fires an event and resets its membrane potential. This firing mechanism emulates how biological neurons emit a spike when they reach their activation threshold. In the context of object detection, this spiking behavior is highly beneficial for detecting transient or fast-moving objects, as they produce dense bursts of events due to rapid brightness changes. This property makes the event rate directly proportional to the speed of the object:
\begin{equation}
    \text{Event Rate} \;\propto\; \text{Object Speed},
    \label{ev_rate}
\end{equation}
    
As a result, fast-moving objects produce dense event bursts, which accumulate quickly and cause the membrane potential to surpass $V_{\mathrm{th}}$.

To localize objects within the event stream after a neuron fires, we extract a bounding box based on the spatial distribution of spiking events. This is achieved by computing the minimum and maximum coordinates of the spiking pixels:

These coordinates represent the bounding limits of the detected object. We then calculate the center of the bounding box to determine the object's approximate position:

\begin{subequations}
\label{eq:center}
\begin{align}
\text{center}_x &= X_{\min} + \Bigl\lfloor\frac{X_{\max} - X_{\min}}{2}\Bigr\rfloor,\\
\text{center}_y &= Y_{\min} + \Bigl\lfloor\frac{Y_{\max} - Y_{\min}}{2}\Bigr\rfloor.
\end{align}
\end{subequations}

This computation yields the center point of the bounding box, providing an efficient method for locating objects within the event frame.

As shown in Figure~\ref{fig:ev_box}, the single-layer SNN is able to localize the target even at greater depths, where event density is lower. We use a $\beta = 0.1$, $V_{th}=1.75$. This was obtained after fine-tuning for the given gate which moves at $4$ m/s. We use a $3\times3$ sized kernel for $W$. See Section \ref{ssec:tracking_results} for further details.

\subsection{\textbf{Physics-Guided Trajectory-Duration Predictor}}
\label{subsec:physics_guided}

\begin{figure}[t]
    \centering
    \includegraphics[width=0.95\columnwidth]{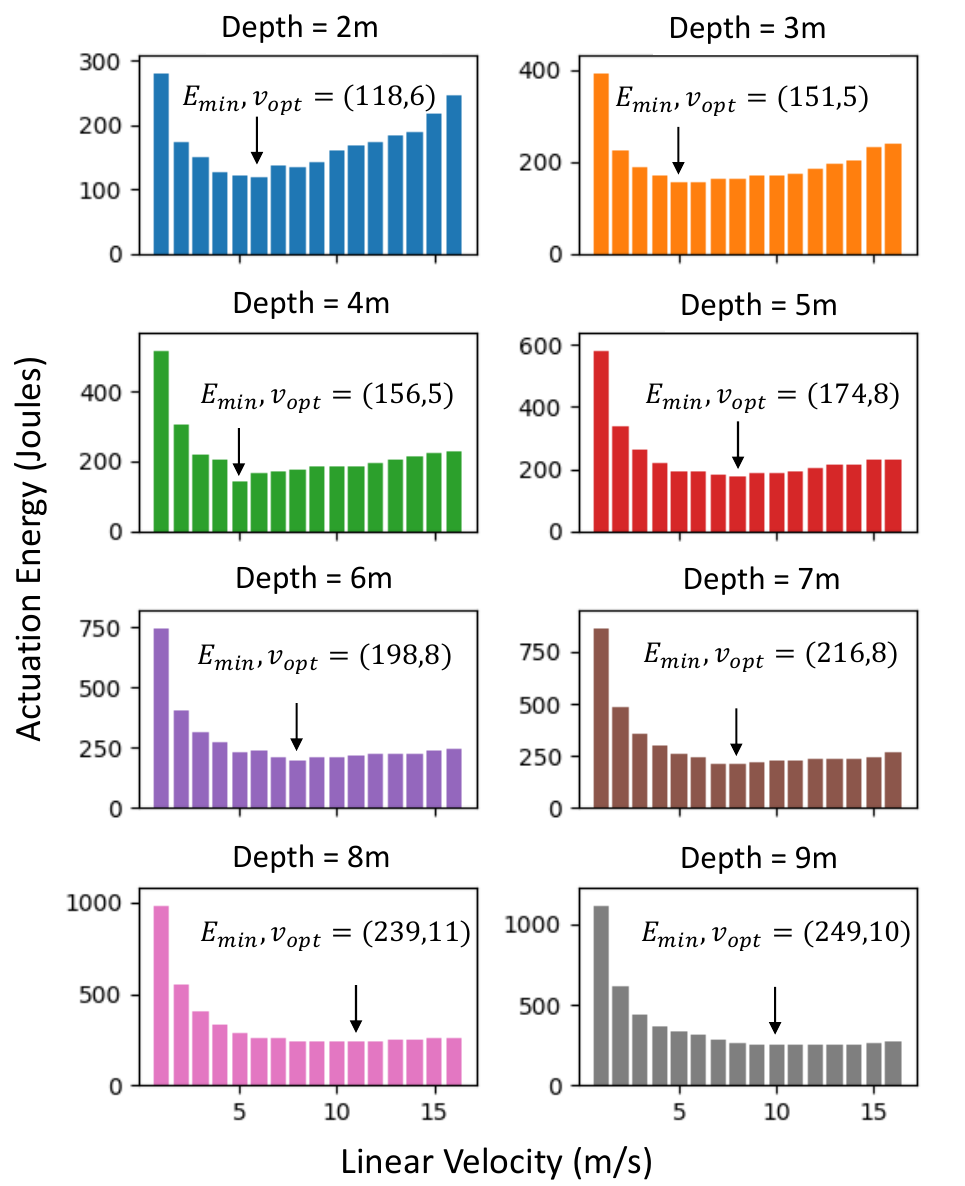}
    \caption{%
        Optimal velocity and energy-consumption patterns. 
        Each depth features a characteristic velocity $v_{\mathrm{opt}}$ 
        that balances time and power usage.
    }
    \label{fig:energy_time}
\end{figure}

We model each quadrotor propeller by:
\begin{equation}
    e(t) \;=\; R\,i(t) + K_E\,\omega(t),
    \label{eq:quadrotor_propeller}
\end{equation}

where $R$ is the winding resistance, $i(t)$ the current, and $\omega(t)$ the rotor speed. The total energy from $t=0$ to $t=T$ is given by:
\begin{equation}
    E(T) \;=\; \int_{0}^{T} \sum_{j=1}^{4} e_j(\tau)\,i_j(\tau)\,d\tau,
    \label{eq:E}
\end{equation}
revealing that flying very slowly or very fast can waste energy. Each depth $d$ therefore has an ideal velocity $v_{\mathrm{opt}}$ minimizing overall consumption, illustrated in Figure~\ref{fig:energy_time}.

A \emph{physics-guided neural network} (PgNN) learns this relationship by approximating \(E(v)\) through polynomial regression for each depth. By fitting a 5th-degree polynomial to the energy-velocity data, the PgNN captures the non-linear dynamics inherent in the system. The optimal velocity \(v_{\mathrm{opt}}\) is determined by finding the velocity at which the derivative of the energy function equals zero:
\begin{equation}
    \frac{dE(v)}{dv} = 0 \quad \Rightarrow \quad v_{\mathrm{opt}} = \arg\min_v E(v),
    \label{eq:optimal_velocity}
\end{equation}
The PgNN predicts \(v^{\mathrm{pred}}\), yielding an approximate flight time:
\begin{equation}
    t_{\mathrm{traj}} 
    \;=\; \frac{d}{v^{\mathrm{pred}}},
    \label{eq:ttraj}
\end{equation}

\begin{table}[ht]
\centering
\caption{PgNN Training Samples: Depth, Velocity, and Physical Constraints obtained by fiiting polynomial curves}
\label{tab:data}
\begin{tabular}{@{}lcc@{}}
\toprule
\textbf{Depth} & \textbf{Velocity} & \textbf{Constraint}                  \\ \midrule
$d_1$          & $v_1$             & $c_{1,1} + 2\,c_{2,1}\,v_1 + \ldots = 0$       \\
$d_2$          & $v_2$             & $c_{1,2} + 2\,c_{2,2}\,v_2 + \ldots = 0$       \\
$\cdots$       & $\cdots$          & $\cdots$                                      \\
$d_n$          & $v_n$             & $c_{1,n} + 2\,c_{2,n}\,v_n + \ldots = 0$       \\ \bottomrule
\end{tabular}
\end{table}

The samples used to train the PgNN are summarized in Table~\ref{tab:data}, which lists various depths $d_n$, their corresponding optimal velocities $v_n$, and the associated constraints derived from the polynomial fits of $E(v)$ and its derivative.

In battery-constrained aerial systems, ensuring energy-efficient flight requires careful consideration of power consumption, \(P(t)\). Power consumption is directly related to the thrust force as:
\begin{equation}
    P(t) = \kappa \|\mathbf{F}_{\text{thrust}}(t)\|^\alpha,
    \label{eq:power_consumption}
\end{equation}
where \(\kappa\) and \(\alpha\) are constants determined by the propeller actuator characteristics of the UAV. The parameter \(\alpha\) governs the non-linearity of the power-thrust relationship and has a significant impact on the energy expenditure profile across varying flight velocities.

\subsection{\textbf{PgNN Loss Function}}

The Physics-Guided Neural Network (PgNN) is trained to predict near-optimal velocities by minimizing a composite loss function:
\begin{equation}
    \mathcal{L}_{\mathrm{PgNN}} = \mathcal{L}_{\mathrm{data}} + \lambda_1 \mathcal{L}_{\mathrm{physics}} + \lambda_2 \mathcal{L}_{\mathrm{energy}},
    \label{eq:pgnn_loss}
\end{equation}
where each term contributes to a specific aspect of the optimization:

\paragraph{Data-Fitting Loss (\(\mathcal{L}_{\mathrm{data}}\))} Ensures that the PgNN predictions align with the ground truth velocity \(v_i^{\mathrm{opt}}\), derived from offline optimization or simulation data, as follows:
\begin{equation}
    \mathcal{L}_{\mathrm{data}} = \frac{1}{N} \sum_{i=1}^N \bigl(v_i^{\mathrm{pred}} - v_i^{\mathrm{opt}}\bigr)^2.
    \label{eq:data_fitting_loss}
\end{equation}
Here, \(v_i^{\mathrm{opt}}\) is linked to the trajectory time \(t_{\mathrm{traj}}\) (as defined in Equation~\eqref{eq:ttraj}), enabling the PgNN to learn velocity predictions that minimize energy consumption while ensuring timely navigation.

\paragraph{Physics Consistency Loss (\(\mathcal{L}_{\mathrm{physics}}\))} Ensures that the PgNN respects the physical laws governing UAV motion:
\begin{equation}
    \mathcal{L}_{\mathrm{physics}} = \frac{1}{N} \sum_{i=1}^N \bigl|\mathbf{x}_i^{\mathrm{pred}} - \mathbf{x}_i^{\mathrm{sim}}\bigr|,
    \label{eq:physics_consistency_loss}
\end{equation}
where \(\mathbf{x}_i^{\mathrm{sim}}\) represents the UAV states predicted by a physics-based simulation model (Equation \ref{eq:dynamics}). The dynamics of the UAV are governed by Equation~\eqref{eq:power_consumption}, which capture energy consumption \(P(t)\) and the time-energy trade-off in the objective function.

\paragraph{Energy Efficiency Loss (\(\mathcal{L}_{\mathrm{energy}}\))} Promotes predictions that minimize energy consumption, calculated using the power consumption equation:
\begin{equation}
    \mathcal{L}_{\mathrm{energy}} = \frac{1}{N} \sum_{i=1}^N P(t_i^{\mathrm{pred}}),
    \label{eq:energy_efficiency_loss}
\end{equation}
where \(P(t)\) is defined in Equation~\eqref{eq:power_consumption}, and its computation integrates over the trajectory time \(t_{\mathrm{traj}}\). This term enforces efficiency by penalizing excessive energy use across predicted trajectories.

The hyperparameters \(\lambda_1\) and \(\lambda_2\) are tuned to balance the contributions of the physics and energy terms, ensuring that the PgNN achieves both accurate predictions and energy-efficient trajectories. By combining these loss components, the PgNN learns velocity predictions that optimize flight time and energy consumption while respecting physical constraints.
In this work, the PgNN is a $3$-layer fully connected multi-layer perceptron with $[64,128,128]$ neurons. Later, in Section \ref{subsec:reg}, we analyze the effect of varying the hyperparameters of our PgNN.

\subsection{\textbf{Symbolic Planning for Moving Gates}}
\label{symbol}

The system integrates a rule-based planner to handle dynamic obstacles such as moving gates as shown in Algorithm 1 in Fig \ref{fig:method}. In scenarios like drone racing, gates often oscillate along the \(y\)-axis within a bounded range of \(\pm L\). Accurate planning requires predicting the gate's future position based on its current motion and the estimated time of arrival (\(t_{\mathrm{traj}}\)).

Using consecutive gate positions \(\{y_1, y_2\}\) recorded at time intervals of \(\delta t\), the gate's velocity is computed as:
\begin{equation}
    v_r = \frac{y_2 - y_1}{\delta t}.
    \label{eq:gate_velocity}
\end{equation}
This velocity, combined with the predicted time of arrival \(t_{\mathrm{traj}}\) from the Physics-Guided Neural Network (PgNN), estimates the gate's future position:
\begin{equation}
    y^* = y_2 + v_r \cdot t_{\mathrm{traj}}.
    \label{eq:gate_position_prediction}
\end{equation}
If the gate is expected to bounce off a boundary during \(t_{\mathrm{traj}}\), its displacement is adjusted to reflect the change in direction:
\begin{equation}
    y^* = 
    \begin{cases} 
    y_2 - L + x, & \text{if the gate bounces to the left}, \\
    y_2 + L - x, & \text{if the gate bounces to the right}.
    \end{cases}
    \label{eq:gate_boundary_adjustment}
\end{equation}
In this way, the proposed framework ensures collision-free navigation while minimizing energy usage.

\begin{table*}[tb]
\centering
\caption{Comparison of tracking methods (SNN, YOLO, R-CNN, and Hough Transform)}
\label{tab:comparison}
\resizebox{0.9\linewidth}{!}{%
\begin{tabular}{lcccc}
\toprule
\textbf{Method} & 
\textbf{Parameter Count} & 
\textbf{Training / Labeling} & 
\textbf{Thresholds} & 
\textbf{Notes} \\
\midrule
\textbf{SNN \cite{nagaraj2023dotie,evplanner,joshi2024}} & 
\begin{tabular}[c]{@{}l@{}}
1 Conv2D \((3\times3)\)\\
9 Spiking Neurons
\end{tabular} & 
Unsupervised & 
\begin{tabular}[c]{@{}l@{}}
Yes (firing \\
threshold)
\end{tabular} &
Lightweight, event-driven \\
\midrule
\textbf{YOLO (CNN-based) \cite{yolo}} & 
\(\sim\!60\text{-}65\)M params & 
\begin{tabular}[c]{@{}l@{}}
Requires large \\
labeled dataset
\end{tabular} &
No &
Popular real-time detector \\
\midrule
\textbf{Faster R-CNN (CNN-based) \cite{rcnn}} & 
\(\sim\!60\text{-}100\)M params & 
\begin{tabular}[c]{@{}l@{}}
Requires large \\
labeled dataset
\end{tabular} &
No &
Region proposals + deep CNN \\
\midrule
\textbf{Hough Transform \cite{hough_t}} & 
\textit{N/A} & 
No (classical CV) & 
Yes (edges, thresholds) &
Handcrafted approach \\
\bottomrule
\end{tabular}%
}
\end{table*}

\begin{figure*}[htbp]
    \centering
    \includegraphics[width=1\textwidth]{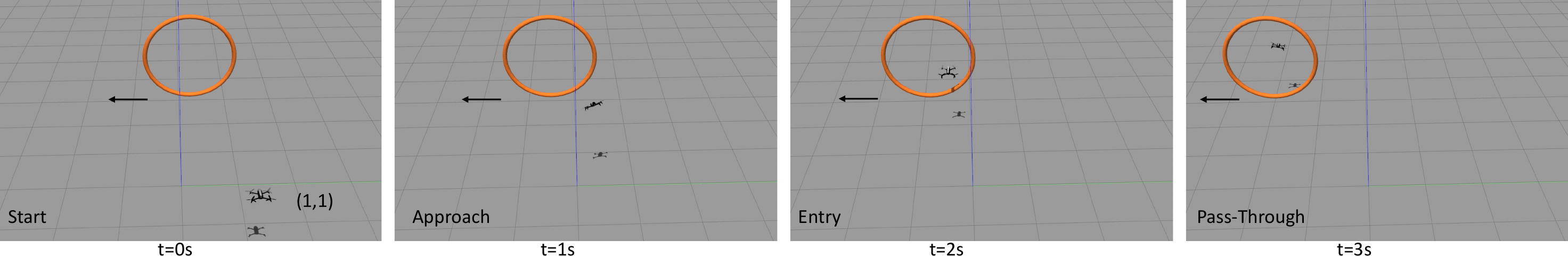} 
    \caption{Drone navigating through a moving gate in Gazebo simulation. The gate moves from right to left, while the drone starts at position \((1, 1)\). The timestamps (`t = 0s`, `t = 1s`, `t = 2s`, `t = 3s`) indicate the time elapsed at each key stage: (a) Start, (b) Approach, (c) Entry, and (d) Pass-Through.}
    \label{fig:drone_navigation_timeline}
\end{figure*}
\begin{figure}[tb]
    \centering
    \includegraphics[width=0.9\linewidth]{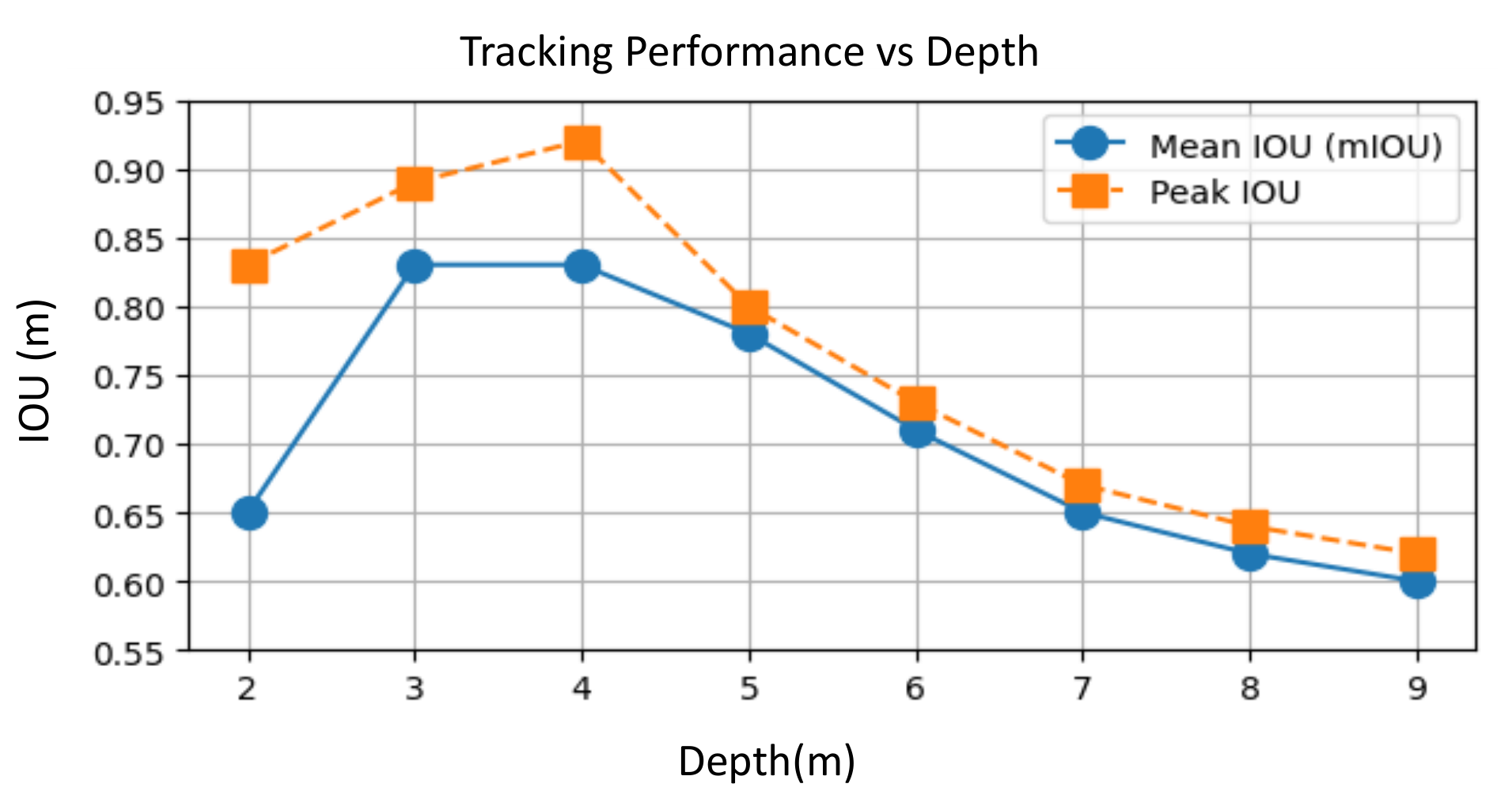}
    \caption{Tracking performance vs.\ depth. The plot compares the mean IoU (mIOU)
    and the peak IoU across distances from 2\,m to 9\,m.}
    \vspace{-2mm}
    \label{fig:iou_ijcnn}
\end{figure}
\begin{figure}[ht]
    \centering
    \includegraphics[width=\linewidth]{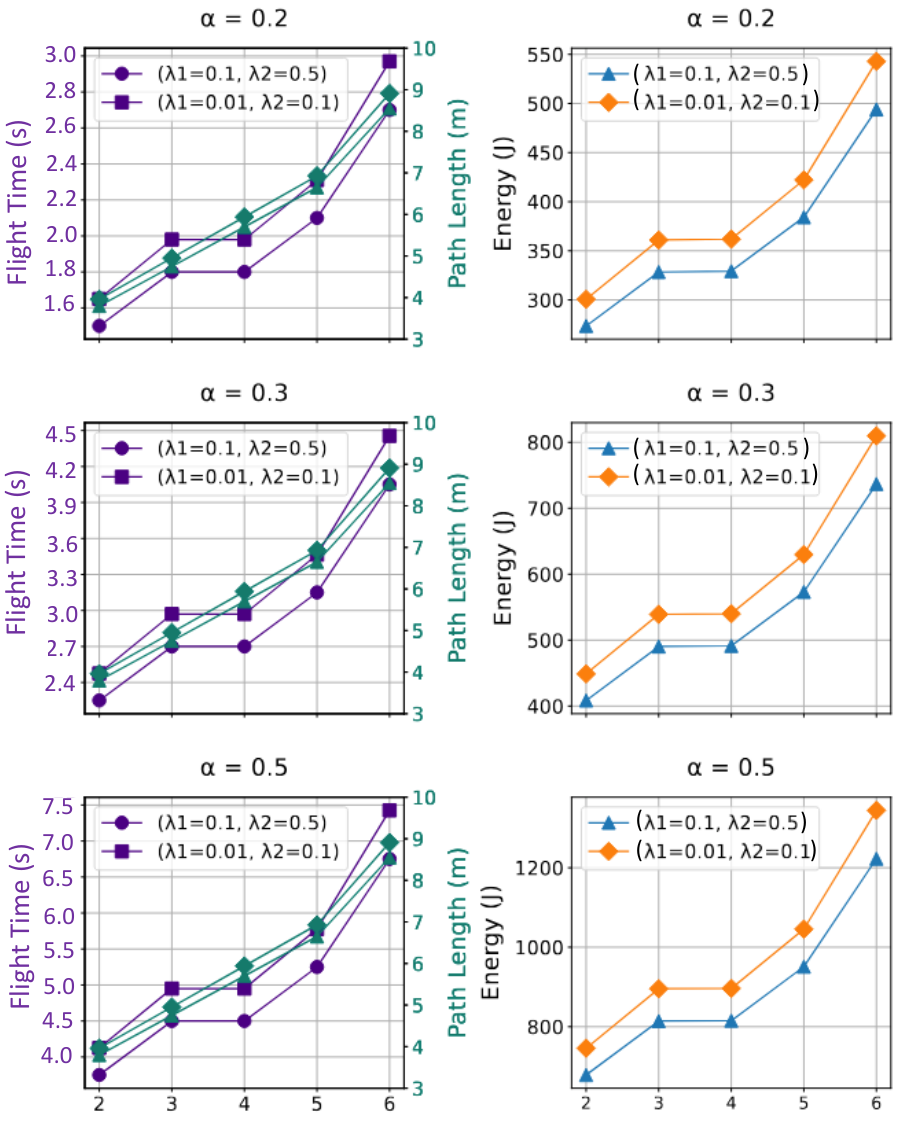}  
    \caption{Comparison of flight time, path length, and dynamic energy across varying depths for different values of $\alpha$ and $\lambda$. The left column displays the combined plots of flight time (in seconds) and path length (in meters), showcasing the differences influenced by the regularization weights $\lambda_1$ and $\lambda_2$. The right column shows the corresponding dynamic energy (in joules) for the same parameter sets. Each row corresponds to a different power model exponent $\alpha$ (0.2, 0.3, and 0.5).}
    \label{fig:flight_energy_plot}  
\end{figure}

\begin{figure*}[!t]
    \centering
    \includegraphics[width=0.95\textwidth]{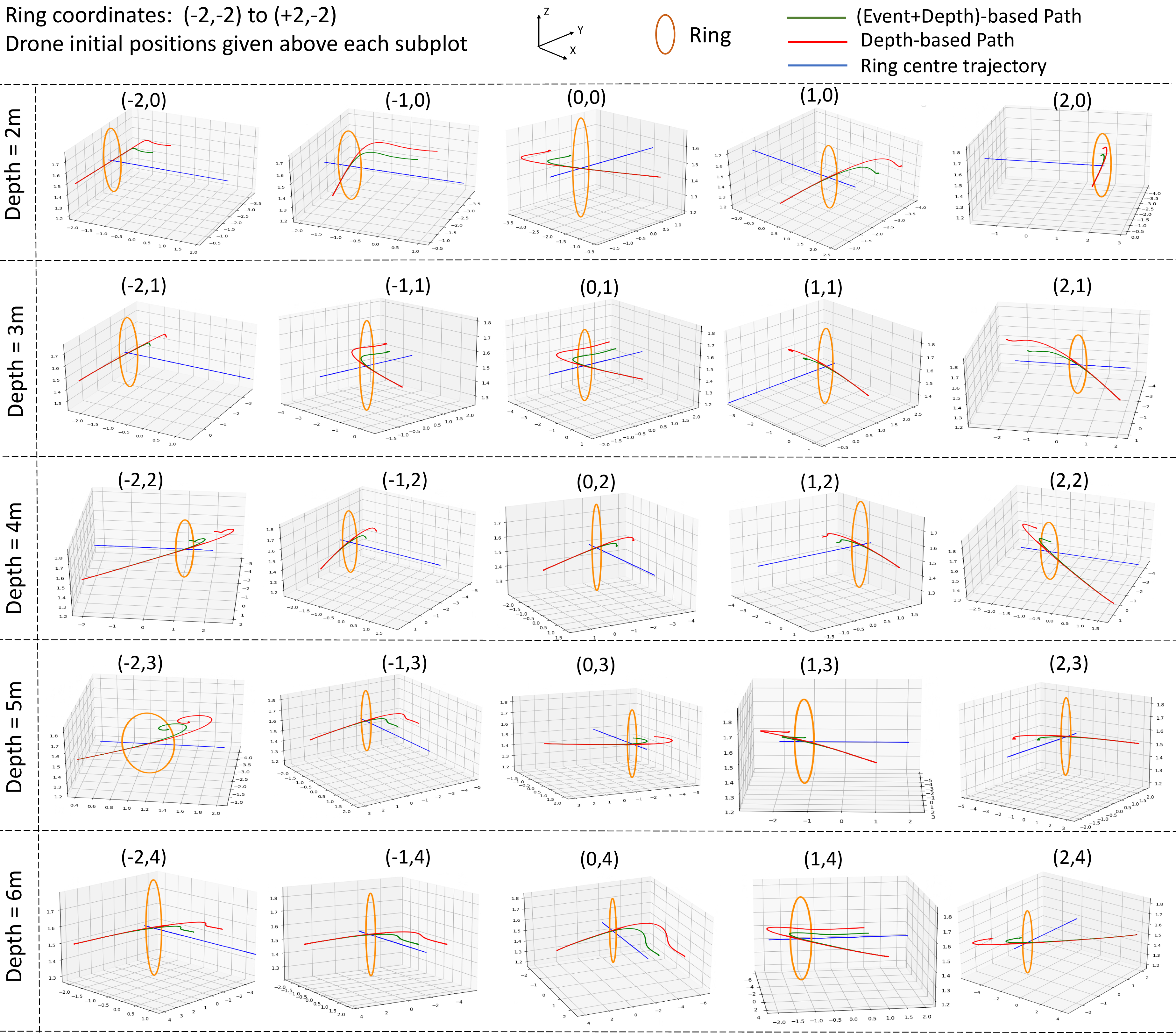}
    \caption{Comparison of Navigation Trajectories: The figure illustrates the navigation paths of a drone using depth-based perception (red trajectory) versus a physics-guided neuromorphic vision-based approach that fuses event and depth sensors (green trajectory). The neuromorphic approach consistently results in shorter, more energy-efficient trajectories across varying initial drone positions and ring depths.}
    \label{fig:3d_navigation_paths}
\end{figure*}
\begin{figure}[t!]
    \centering
    \includegraphics[width=\linewidth, height = 3.3cm]
    {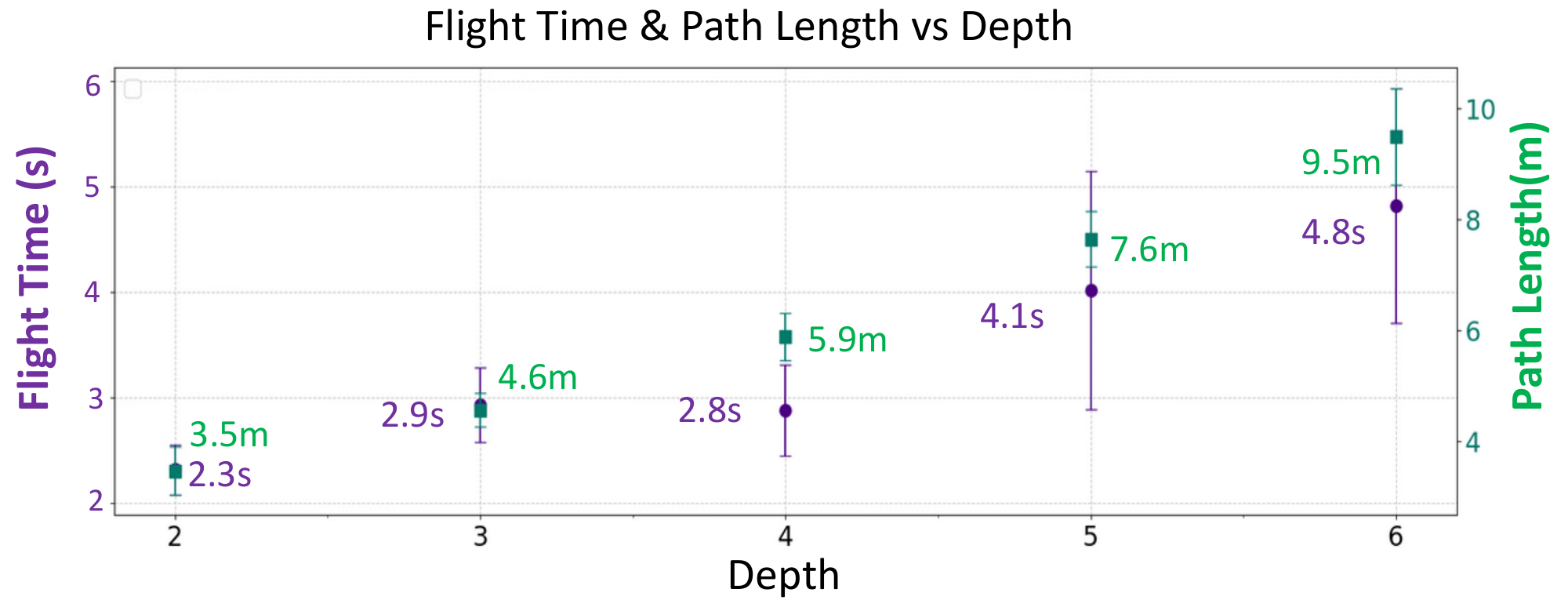}
    \caption{Flight time (left y-axis) and path length (right-y axis) as functions of depth for different initial drone positions for our physics-guided neuromorphic approach. The left y-axis shows the flight time in seconds, and the right y-axis shows the corresponding path length in meters. The data is grouped by depth values, with distinct markers indicating measurements for each combination of flight time and path length.}
    \label{fig:flight_path_depth}
\end{figure}

\noindent{\textbf{Neurosymbolic Integration for Navigation}}: The planner incorporates inference results from multiple neural network modules. The event-based Spiking Neural Network (SNN) detects and tracks the moving gate, providing bounding box coordinates \((y_1, y_2)\). The Physics-Guided Neural Network (PgNN) predicts the trajectory time \(t_{\mathrm{traj}}\) while optimizing for energy efficiency. These outputs are fused and processed through logical rules, forming a neurosymbolic framework that combines data-driven predictions with rule-based reasoning. This integration enables the system to account for boundary dynamics, predict gate behavior, and compute the gate’s future position \(y^*\). By leveraging this neurosymbolic approach, the system achieves efficient and safe navigation through moving gates. Please note, this work only considers local reactive planning, necessary when there are changes observed in the immediate surrounding environment of the robotic agent (drone in this work).

\section{Experimental Results}

\subsection{\textbf{Neuromorphic Vision-Based Tracking}}
\label{ssec:tracking_results}
Figure \ref{fig:ev_box} shows neuromorphic event output and a bounding box 
around a moving gate (left to right). Although event density decreases
with increasing depth, the SNN continues to isolate and track the gate
in real time. Correspondingly, as illustrated in Figure~\ref{fig:iou_ijcnn}, both the mIOU and the peak IOU gradually decline as the distance between the event-based camera and the moving gate increases. At shorter ranges (2--4\,m), the system achieves a mean IOU between 0.78 and 0.83, with a peak IOU above 0.90 at 4\,m. This high accuracy stems from dense event bursts generated by the gate’s motion, allowing the spiking neural network (SNN) to consistently localize the target with minimal false positives.

However, at larger depths (beyond 6,m), the IOU values drop slightly (mIOU: 0.60--0.65; peak IOU: 0.62--0.73) due to sparser events and weaker contrast changes. Despite this, the neuromorphic detection pipeline maintains robust tracking across a wide range of depths, demonstrating its suitability for low-latency object detection in dynamic flight scenarios.

As summarized in Table~\ref{tab:comparison}, our shallow SNN-based approach has an extremely low parameter count (one \(3\times 3\) convolutional layer and 9 spiking neurons). It operates without a large labeled dataset but still requires threshold tuning (e.g., spiking neuron firing thresholds). Meanwhile, YOLO and R-CNN each demand tens of millions of parameters, along with substantial labeled training data. Finally, the classical Hough Transform method does not rely on deep learning but involves hand-tuned thresholds for edge detection. Our SNN maintains a balance of minimal parameter overhead, unsupervised learning, while offering the benefits of neuromorphic perception using SNN-based dynamic vision sensors.

\subsection{\textbf{Sensitivity Analysis for Physics-based Regularization}}
\label{subsec:reg}
To examine how the power-model exponent \(\alpha\) (Equation~\ref{eq:power_consumption}) and the regularization weights \(\lambda_1\) and \(\lambda_2\) (Equation~\ref{eq:pgnn_loss}) jointly shape flight efficiency, we conducted a series of simulations at varying ring depths and initial drone positions. Figure~\ref{fig:flight_energy_plot} presents the resulting flight time, path length, and dynamic energy for three representative \(\alpha\) values (0.2, 0.3, and 0.5). Each row corresponds to one exponent, while different lines indicate distinct \(\lambda_1,\lambda_2\) settings. The left column combines flight time (in seconds) and path length (in meters), highlighting how \(\alpha\) and \(\lambda\)-values modify navigation efficiency. The right column displays the corresponding dynamic energy (in joules), illustrating the interplay between control regularization and power consumption.

As shown in the left-column plots of Figure~\ref{fig:flight_energy_plot}, increasing the ring depth results in longer flight times and paths, especially at higher \(\alpha\). This effect is magnified when \(\lambda\)-values are small, since weaker regularization allows more aggressive maneuvers, potentially increasing path deviations. Conversely, higher \(\lambda_1,\lambda_2\) steer the flight toward smoother trajectories, dampening sudden control changes and thereby reducing energy spikes (right-column plots). 
Note that, the top row ($\alpha = 0.2$) reflects parameters tuned to our Bebop simulation, while the middle and bottom rows ($\alpha = 0.3$ and $\alpha = 0.5$) illustrate hypothetical regimes. This highlights the sensitivity of the power‐consumption model to $\alpha$; in real‐world applications, calibrating $\alpha$ to the specific drone is crucial for accurately capturing its flight dynamics and energy demands. 

Also, note that the energies reported for inference during the actual flights are higher than the ones presented in Figure \ref{fig:energy_time}, as the actual flight paths are longer than the straight line flight distances covered (no moving gate there) while collecting training data (Table \ref{tab:data}).

 In the following subsection, we expand our analysis to examine how these parameter choices translate into navigation performance across a range of drone starting positions and ring depths.

\subsection{\textbf{Neuromorphic Navigation for different Drone Positions }}
Figure~\ref{fig:drone_navigation_timeline} illustrates the drone's navigation through a moving ring, with timestamps marking key stages from start ($t = 0s$) to pass-through ($t = 3s$), highlighting the smooth progression despite the dynamic nature of the ring.

Figure~\ref{fig:3d_navigation_paths} illustrates the comparison of navigation trajectories for a drone navigating through a ring from various initial positions \((x, y, z)\) and ring depths ranging from 2~m to 6~m. The figure highlights the performance of the depth-based perception approach (red trajectory) versus the physics-guided neuromorphic vision-based approach (green trajectory), which combines event-based and depth-based sensor inputs. The trajectory of the ring's center is depicted in orange. We observe the following:
\begin{itemize}
    \item \textbf{Energy-Efficient Trajectories:} The neuromorphic vision-based approach produces shorter paths, such as achieving a path length of 3.5~m at a 2~m depth, compared to 4.6~m for an off-center start.
    
    \item \textbf{Accuracy and Responsiveness:} When starting from positions \((-2, y, z)\) or \((+2, y, z)\), the neuromorphic approach maintains a shorter flight time of around 2.9~s at a depth of 3~m, compared to longer times exceeding 4.1~s for off-center starts at greater depths.
    
    \item \textbf{Trajectory Deviation:} Depth-based perception results in longer paths, such as 7.6~m at a 5~m depth, compared to the neuromorphic approach’s 5.9~m for the same conditions, demonstrating the effect of initial offsets.
    
    \item \textbf{Impact of Depth:} The neuromorphic approach remains efficient with path lengths ranging from 3.5~m (at 2~m) to 9.5~m (at 6~m), while the depth-based method increases path length at greater depths, up to 10~m or more.
\end{itemize}

Figure~\ref{fig:flight_path_depth} further illustrates the dependence of flight time and path length on ring depth for various initial drone positions. Here, the left y-axis measures flight time (in seconds), while the right y-axis tracks path length (in meters). Distinct markers indicate different initial offsets \((x, y)\), showing that off-center starts tend to increase the required flight time and distance. For instance, at a depth of 3~m, the flight time increases from 2.3~s (centered start) to 2.9~s (off-center start), with the corresponding path length increasing from 3.5~m to 4.6~m. At a depth of 5~m, the off-center start results in a flight time of 4.1~s and a path length of 7.6~m. These results highlight the importance of appropriate tuning of \(\alpha\) and \(\lambda\)-values to balance energy conservation against the need for precise maneuvers. Overall, the physics-guided neuromorphic approach reduces average flight time by approximately 20\% and path length by around 15\% compared to the depth-based method, demonstrating its potential for more efficient autonomous aerial navigation.

\section{Conclusion}
This work presents a novel framework for energy-efficient autonomous aerial navigation that leverages neuromorphic event-based vision and physics-guided planning. By integrating dynamic vision sensors (DVS) with spiking neural networks (SNNs) and physics-guided neural networks (PgNNs), the system achieves real-time responsiveness and energy optimization. Experimental results demonstrate that the proposed approach produces shorter, energy-efficient trajectories with high accuracy and adaptability across varying initial positions and depths. Notably, the fusion of asynchronous event data with depth information enhances trajectory planning by enabling robust navigation through dynamic environments, such as moving gates. The neuromorphic design outperforms traditional depth-based methods due to its low-latency processing and efficient handling of sparse data, making it particularly advantageous for scenarios for real-time decision-making.

The proposed framework is also a neurosymbolic integration that combines event-driven neuromorphic computing, physics-based AI, symbolic logical reasoning, and classical planning. The neuromorphic SNN module processes asynchronous event data for low-latency perception, while the PgNN embeds physical constraints to generate energy-optimal trajectories that align with real-world dynamics. The rule-based planner introduces symbolic reasoning to anticipate obstacle movements and guide navigation decisions. This synergy between neural and symbolic components allows the system to handle dynamic, real-time scenarios while maintaining interpretable and physically consistent control. By uniting data-driven perception with rule-based reasoning, the framework achieves adaptable decision-making, highlighting possible benefits of neurosymbolic design for autonomous systems.

\section{Acknowledgement}
This work was supported by the Center for the Co-Design of Cognitive Systems (CoCoSys), a center in JUMP 2.0, an SRC program sponsored by DARPA.

\bibliographystyle{IEEEtran}
\bibliography{conference_101719}


\end{document}